\documentclass[11pt,a4paper]{article}
\usepackage[hyperref]{acl2021}
\usepackage{times}
\usepackage{latexsym}

\usepackage{array}
\usepackage{graphicx}
\usepackage{tabularx,tabulary}
\usepackage{url}
\usepackage{nicefrac}
\usepackage{calc}
\usepackage{xcolor}
\usepackage{graphicx}
\usepackage{multirow}

\usepackage{amsthm}
\usepackage{amssymb}
\usepackage{amsmath}
\usepackage{amsfonts}
\usepackage[toc]{appendix}
\usepackage{color, colortbl}
\usepackage{adjustbox}
\usepackage{tabularx}
\usepackage{booktabs}
\usepackage{subcaption}
\usepackage{makecell}
\usepackage{arydshln}
\usepackage{tikz}
\usepackage{enumitem}
\usepackage{amsmath}

\usepackage{microtype}

\aclfinalcopy 

\usepackage{amsmath,amsfonts,bm}

\def\eqref#1{equation~\ref{#1}}

\def\1{\bm{1}}

\def\vb{{\bm{b}}}

\def\vh{{\bm{h}}}

\def\vw{{\bm{w}}}

\def\mS{{\bm{S}}}

\def\mW{{\bm{W}}}

\DeclareMathAlphabet{\mathsfit}{\encodingdefault}{\sfdefault}{m}{sl}
\SetMathAlphabet{\mathsfit}{bold}{\encodingdefault}{\sfdefault}{bx}{n}

\title{Schema-Guided Paradigm for Zero-Shot Dialog} 
\author{Shikib Mehri \and Maxine Eskenazi \\
  Language Technologies Institute, Carnegie Mellon University \\
  \texttt{\{amehri,max\}@cs.cmu.edu}}

\date{}

\begin{document}
\maketitle
\begin{abstract}
Developing mechanisms that flexibly adapt dialog systems to unseen tasks and domains is a major challenge in dialog research. Neural models implicitly memorize task-specific dialog policies from the training data. We posit that this implicit memorization has precluded zero-shot transfer learning. To this end, we leverage the \textbf{schema-guided paradigm}, wherein the task-specific dialog policy is explicitly provided to the model. We introduce the Schema Attention Model (\textsc{sam}) and improved schema representations for the STAR corpus. \textsc{sam} obtains significant improvement in zero-shot settings, with a \textbf{+22} $\mathbf{F_1}$ score improvement over prior work. These results validate the feasibility of zero-shot generalizability in dialog. Ablation experiments are also presented to demonstrate the efficacy of \textsc{sam}.
\end{abstract}

\section{Introduction}

Task-oriented dialog systems aim to satisfy user goals pertaining to certain tasks, such as booking flights \citep{hemphill-etal-1990-atis}, providing transit information \citep{raux2005let}, or acting as a tour guide \citep{budzianowski2018multiwoz}. Neural models for task-oriented dialog have become the dominant paradigm \citep{williams2016end,wen2016network,zhao2017generative}. These data-driven approaches can potentially learn complex patterns from large dialog corpora without hand-crafted rules. However, the resulting models struggle to generalize beyond the training data and under-perform on unseen dialog tasks and domains \citep{zhao2018zero,rastogi2020towards}.

A long-standing challenge in dialog  research is to flexibly adapt systems to new dialog domains and tasks \citep{zhao2018zero,mosig2020star}. Consider a system that has been trained to handle several different tasks (e.g., restaurant reservations, ride booking, weather, etc.). How can this dialog system be \textit{extended to handle a new task} (e.g., hotel booking), without collecting additional data? This paper tackles this challenge and aims to address the problem of zero-shot generalization using the \textbf{schema-guided paradigm}. 

\begin{figure}
    \centering
    \includegraphics[width=0.5\textwidth]{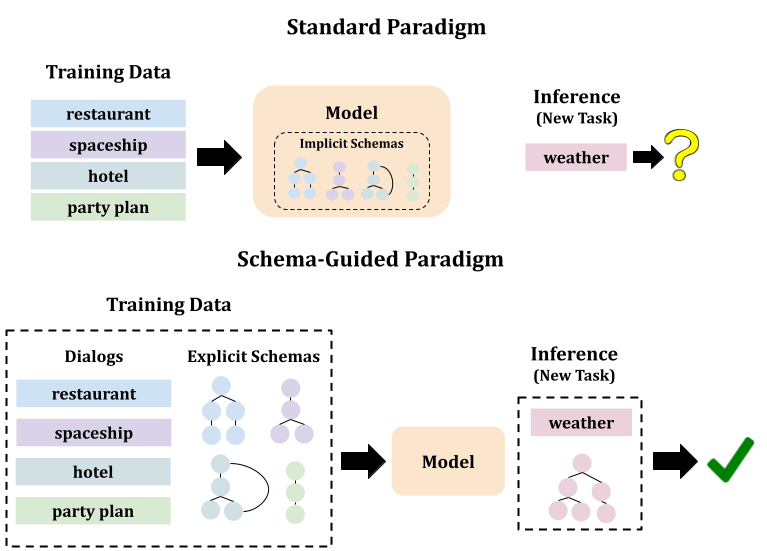}
    \caption{In the standard paradigm, data driven models implicitly learn the task-specific dialog policies (i.e., schemas). This precludes generalization to an unseen task at inference time. In contrast, in the schema-guided paradigm, dialog policy is explicitly provided to the model through a schema graph. At inference time, the model is given the schema for the new task and can therefore generalize in a zero-shot setting. }
    \label{fig:schema_paradigm}
\end{figure}
The advent of large-scale pre-training \citep{devlin-etal-2019-bert,radford2019language} has led to significant progress in domain adaptation across areas in NLP, including natural language understanding \citep{wang2018glue,wang2019superglue}, open-domain dialog \citep{zhang-etal-2020-dialogpt,adiwardana2020towards} and language understanding for task-oriented dialog \citep{wu2019transferable,mehri2020dialoglue}. Generalization in end-to-end task-oriented dialog has proven to be significantly more difficult, particularly in zero-shot settings where there is no training data \citep{mosig2020star}. We posit that it is inherently difficult to generalize to unseen dialog tasks because of the \textbf{dialog policy}.

Traditionally, an end-to-end dialog system must perform three distinct tasks. First, it must understand the dialog history and identify any relevant user intents or slots. Next, it must decide on the appropriate system action, according to a task-specific dialog policy. Finally, it must generate a natural language utterance corresponding to the system action. In a pipeline dialog system, these three steps are performed by the NLU, DM and NLG respectively \citep{jurafsky2000speech}. Prior work has exhibited generalizability in language understanding and, to a lesser extent, in language generation. However for end-to-end dialog, the task-specific dialog policy inherently precludes zero-shot generalization. An end-to-end dialog model trained on several tasks, will \textit{implicitly} learn the dialog policies from the data. However, when generalizing to a new task in a zero-shot setting, the model has no knowledge of the dialog policy for the \textit{new} task. 

To address the difficulty of generalizing to new task-specific dialog policies and in order to facilitate zero-shot generalization, we present the \textbf{schema-guided paradigm}. Generally, end-to-end neural models implicitly learn the task-specific dialog policies from large corpora. In contrast, in the schema-guided paradigm, we explicitly provide the task-specific dialog policies to the model in the form of a \textbf{schema graph}. The schema graphs define the system's behavior for a specific task (e.g., when the user provides the reservation time, ask them for the number of people). When transferring to an unseen task, the corresponding schema graph is explicitly provided to the model. This enables language understanding and the dialog policy to be decoupled. The model no longer needs to implicitly memorize the task-specific policies from the training data. Instead, the model learns to interpret the dialog history and align it to the schema graph. As such, when transferring to a new task, the schema graph serves as an inductive bias that provides the model with the task-specific dialog policy.

To address the challenge of zero-shot transfer learning, \citet{mosig2020star} presented the STAR corpus and several baseline experiments. We extend their baselines for the task of \textit{next action prediction}. We introduce the \textbf{Schema Attention Model} (\textsc{sam}) and thorough schema representations for the 24 different tasks in the STAR dataset. \textsc{sam} obtains a \textbf{+22 $\mathbf{F_1}$} score improvement over baseline approaches in the zero-shot setting, validating the schema-guided paradigm and demonstrating the feasibility of zero-shot generalization for task-oriented dialog. Our code and model checkpoints are open-sourced and be found at \textbf{\url{https://github.com/shikib/schema_attention_model}}.

\section{Related Work}

\subsection{Zero-Shot Dialog}

Zero-shot transfer learning has been of interest to the dialog research community. Many approaches have been proposed for zero-shot adaptation of specific dialog components. \citet{chen2016zero} present a zero-shot approach for learning embeddings for unseen intents. \citet{bapna2017towards} show that slot names and descriptions can be leveraged to implicitly align slots across domains and achieve better cross-domain generalization. \citet{wu2019transferable} similarly use slot names, in combination with a generative model for state tracking, to obtain strong zero-shot results. \citet{shah2019robust} leverage examples for zero-shot slot filling. Generally, approaches for zero-shot generalizability leverage the similarity across domains (e.g., \textit{restaurant-area} and \textit{hotel-area} are conceptually similar). The advent of large-scale pre-training \citep{devlin-etal-2019-bert,radford2019language} allows for language understanding across dissimilar domains. \citet{rastogi2020schema} address zero-shot domain adaptation in state tracking by leveraging BERT \citep{devlin-etal-2019-bert} with a domain-specific API specification.

\citet{zhao2018zero} present an approach for zero-shot end-to-end dialog. They leverage the Action Matching framework to learn a cross-domain latent action space. \citet{qian2019domain}  use model-agnostic meta learning to attain stronger results in zero-shot dialog. Both these approaches rely on additional annotations, which make them unsuitable for the STAR corpus. While there is a significant amount of work in zero-shot generalizability for language understanding, there is considerably less research in adaptation for end-to-end dialog\footnote{While we focus on next action prediction, in the STAR dataset it is trivial to go from a system action to a natural language response and as such we consider our task to be end-to-end dialog.}. This is in part because of the difficulty of generalizing to unseen task-specific policies. To this end, \citet{mosig2020star} presented STAR, a corpus consisting of 24 different dialog tasks, and several baseline models for zero-shot adaptation on STAR. The results in this paper significantly outperform the baselines introduced by \citet{mosig2020star} as we leverage the schema-guided paradigm for zero-shot generalizability in dialog.

\subsection{Schema-Guided Paradigm}

Plan-based dialog systems \citep{ferguson1998trips,rich1998collagen,bohus2009ravenclaw} reason about user intent, in the context of a \textit{dialog plan}. RavenClaw \citep{bohus2009ravenclaw} consists of a task specification that defines the behavior of a system depending on various user actions. Plan-based dialog systems decouple the task-specific dialog policy from the task-agnostic components of the system. This allows a system to be extended to a new task by updating the task specification. The schema-guided paradigm shares a similar motivation, and aims to disentangle the dialog policy in neural, data-driven dialog systems. 

Several approaches have been presented to discover dialog structure graphs (similar to the schemas in this paper) from data in an unsupervised manner \citep{shi2019unsupervised,qiu2020structured,xu2020discovering,hu2019gsn}. These approaches have been used to enhance generation for open-domain dialog \citep{qiu2020structured,hu2019gsn}. To the best of our knowledge, these dialog structures have neither been used for generation in task-oriented dialog nor in zero-shot settings. While our schemas are similar to these structure graphs, they are hand-crafted similar to those in plan-based dialog systems. Future work may extend our work by leveraging unsupervised structure graph discovery as an alternative to hand-crafted schemas.

\section{Task Definition}

We address the problem of transferring dialog models to unseen tasks and domains \citep{zhao2018zero}. This problem is especially important in real world settings. It is impossible to preconceive every dialog task that users may need (e.g., a COVID-19 information dialog system). Furthermore, collecting new dialog data for each new task is inherently unscalable \citep{rastogi2020towards}. While rule-based/pipeline dialog systems may be easier to extend to new tasks \citep{bohus2009ravenclaw}, there is a tradeoff between the adaptability of non-neural systems and the performance of neural models.

\subsection{STAR Dataset}

The STAR dataset \citep{mosig2020star} was collected for the purpose of studying transfer learning in dialog. The dataset spans 24 different tasks in 13 different domains (e.g., the restaurant domain has \textit{`restaurant-search'} and \textit{`restaurant-reservations'}). The data collection procedure was designed to reduce ambiguity in the system responses and make system actions deterministic. As such, Amazon Mechanical Turk (AMT) workers were given a flow chart diagram for each task. This flow chart defined the task, including the order in which questions should be asked (e.g., ask date before city), how to respond to various user responses and how to query a database. Additionally, in order to minimize variance in the responses from the wizard, \citet{mosig2020star} incorporate a \textit{suggestions module} during data collection. This module maps the wizard utterance to the closest pre-written response (e.g., \textit{`Give me your name'} $\rightarrow$ \textit{`What is your name?'}). In some cases, it is not possible for the AMT worker to use the suggestions module. Nonetheless, the module increases the consistency of the system actions. 

\citet{mosig2020star} present baseline results on the tasks of next action prediction and response generation. The present paper focuses on \textit{next action prediction}. The objective of next action prediction is to predict the correct system action conditioned on the dialog history. Since there is a one-to-one mapping between system actions and corresponding natural language responses, the primary challenge in extending a next action prediction model to response generation resides in learning to accurately fill in the response templates (e.g., \textit{`Your reservation is confirmed for \{date\}'}).

The STAR dataset consists of three different types of dialogs: (1) \textit{happy} single-task dialogs, (2) \textit{unhappy} single-task dialogs and (3) \textit{multi-task} dialogs. Here, \textit{happy} refers to dialogs where the users are cooperative and complete the task. In contrast, \textit{unhappy} dialogs consist of uncooperative users that may change the subject, engage in irrelevant chit-chat and otherwise aim to push the system beyond its capabilities. Since our primary objective is to address zero-shot transfer, we only consider the \textit{happy} single-task dialogs. There are 1537 happy single-task dialogs and 10,364 turns. 

\subsection{Zero-Shot Setting}

In the STAR dataset, there are 23 dialog tasks (13 domains) with \textit{happy} single-task dialogs. We perform two types of transfer learning experiments: task transfer and domain transfer. In task transfer, a model is trained on $n-1$ tasks (i.e., 22) and evaluated on the last one. This is repeated for each of the 23 tasks. For domain transfer, a model is trained on $n-1$ domains (i.e., 12) and evaluated on the last one. In task transfer, there may be some overlap between the training and testing, for example, the domain-specific terminology. In contrast, in domain transfer there is very limited overlap. When the model is tasked with generalizing to the restaurant domain, it has seen nothing related to restaurants during training.

In both of these settings, the model is aware of which task it is being evaluated on, meaning that it can leverage a task specification (e.g., schema) for the new task. This experimental design resembles a real-world setting where a system developer would be aware of the new task. For example, if a developer wanted to extend a dialog system to handle a COVID-19 related question, they would be able to create a new task specification. As such, our goal is to develop a model that can generalize to an unseen task conditioned on a task specification. 

\section{Methods}

In order to enable zero-shot transfer to new dialog tasks and domains, the Schema Attention Model (\textsc{sam}) is introduced. It leverages an external dialog policy representation (i.e., the schema) to predict the next system action. This section begins by describing the baseline model for the task of next action prediction. Next, the schema-guided paradigm is introduced (Figure \ref{fig:schema_paradigm}). It includes a graph-based representation of the task-specific schema and \textsc{sam}, a model that identifies the next system action by attending to a task-specific schema representation.

\subsection{Baseline}

This section describes the baseline model proposed by \citet{mosig2020star}. Given an arbitrary language encoder, denoted as $\mathcal{F}$, the baseline model obtains a vector representation of the dialog history, $c$. This representation is then passed through a softmax layer to obtain a probability distribution over the actions. 

\begin{gather}
    \vh = \mathcal{F}( c ) \\
    P_{\text{clf}} = \text{softmax} ( \mW \vh^T + \vb ) 
\end{gather}

Throughout this paper, BERT-base \citep{devlin-etal-2019-bert} is used as the language encoder. 

\subsection{Schema-Guided Paradigm}

Our baseline model simultaneously needs to (1) interpret the dialog context and identify the relevant intents and slots, and (2) learn the task-specific dialog policies (i.e., if the user wants the weather, ask the city) for the different tasks in the training data. This model is incapable of generalizing to a new task in a zero-shot setting, as it would lack knowledge of the task-specific policy for the new task. To mitigate this problem and to enable zero-shot task transfer, we present the schema-guided paradigm which decouples the task-specific dialog policy from the language understanding.

An example is shown in Figure \ref{fig:schema_paradigm}: the schema-guided paradigm decouples the the dialog policy from language understanding by explicitly providing task-specific schema graphs as input to the model. These schema graphs serve as complete representations of the dialog policy for a given task. Therefore, while the baseline needs to implicitly learn the dialog policies, a schema-guided model instead learns to leverage the explicit schema graphs. As such, a schema-guided model can generalize to a new task as long as it is provided with the corresponding schema graph. 

In this paradigm, the role of the model is to interpret a dialog context and align it to the explicit schema graph. The role of the schema graph is to determine the next action according to the dialog policy. In this manner, the language encoder is being trained for the task of sentence similarity. With the help of pre-trained models, language understanding in a schema-guided paradigm can be considered to be task-agnostic. By decoupling the task-agnostic language understanding and the task-specific dialog policy, the schema-guided paradigm better facilitates zero-shot transfer learning. 

The schema-guided paradigm consists of the representation of the schema graph, and a neural model which interprets the dialog context and aligns it to the schema graph.

\subsubsection{Schema Representation}

In the schema-guided paradigm, the schema representation is the task-specific dialog policy. To ensure the efficacy and robustness of the  dialog system, it is important that the schema representation be complete and informative. In the case of ambiguity or incompleteness in the schema representation, the next action will fail to be correctly predicted, regardless of the strength of the model. The schema representations are manually constructed for every task. In the schema-guided paradigm, to transfer to a new task, a system developer would simply need to construct a new schema representation.

\citet{mosig2020star} propose a baseline schema representation wherein the nodes of the graph correspond to system actions and database states. There are nodes for user states only in situations where the system behavior differs depending on the user's actions (e.g., \textit{`Yes'} $\rightarrow$ \textit{ask-time}, \textit{`No'} $\rightarrow$ \textit{ask-date}). The consequence of this representation is that when the model aligns the dialog history to the schema, it largely relies on the system utterances. However, this representation fails to account for realistic user behavior and therefore yields only marginal improvement over the baseline.

Specifically, users will often provide information out of turn (e.g., \textit{System: `Where would you like to go?' $\rightarrow$ User: `Leaving from the airport and going downtown'}). In this example, it is difficult for the model to realize that the question \textit{System: `Where are you leaving from?'} has also been answered and therefore should not be the next system action. Users can also ignore the system utterance (e.g., \textit{System: `Where would you like to go?'} $\rightarrow$ \textit{User: `Actually, what's the weather?'}). It is thus ineffective to represent dialog policy only in terms of the system utterances. To this end, we extend the schema representation by incorporating user utterances into the schema graph. 

\begin{figure}
    \centering
    \includegraphics[width=0.5\textwidth]{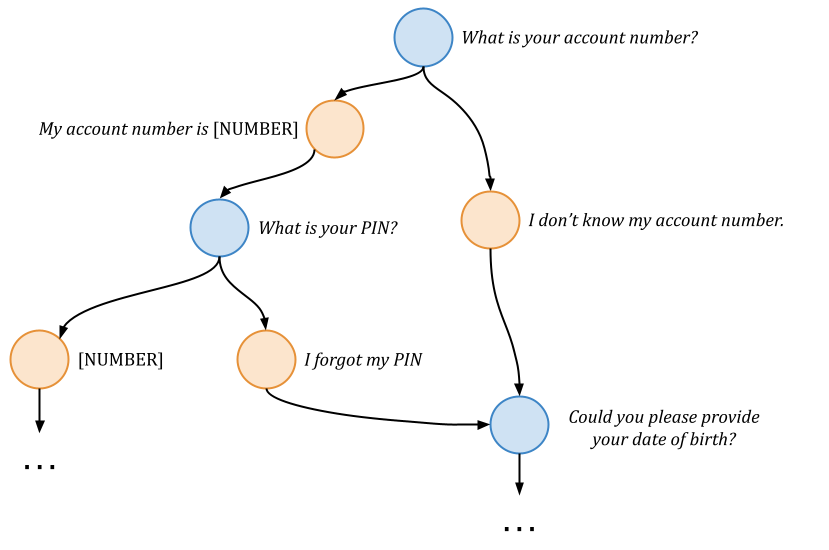}
    \caption{A section of the task-specific schema graph for the \textit{bank-balance} task. The system must authenticate the user with their account number and PIN. However, if the user has forgotten either of these, it must ask backup security questions. The blue nodes correspond to system actions and the yellow nodes denote user utterances.}
    \label{fig:schema_graph}
\end{figure}

As shown in Figure \ref{fig:schema_graph}, our schema graph incorporates nodes corresponding to user utterances. As such, if a user provides information out of turn or changes the subject, our model will be able to effectively align the dialog to the schema. To account for variance in the user utterances, future work could extend this schema representation to include multiple variations of a given user utterance. However, as the schema graphs are manually constructed for every task, there is a trade-off between manual effort and efficacy\footnote{Constructing the schema graphs is not particularly labor-intensive. It took the first author between 15 and 45 minutes to create each schema graph, depending on the complexity of the task.}.

The schema graph has several noteworthy properties. First, the system actions are consistently deterministic. Nodes corresponding to a database response or to a user utterance will always have a single outgoing edge to a system response node. Furthermore, such nodes will also have a single incoming edge from a system response node. For a given user/database node, $u$, we denote the previous system response node as $\textbf{prev}(u)$ and the following system response as $\textbf{next}(u)$. Each node has some text associated with it, denoted as $\textbf{text}(u)$. This text is a template for either a system utterance, database response or user utterance. System nodes will also have an associated system action, $\textbf{act}(u)$. There is a one-to-one mapping between the system actions and the system response templates.

\subsubsection{Schema Attention Model}

In the schema-guided paradigm, the role of the model is to understand the dialog history and align it to the schema representation. We introduce the \textbf{Schema Attention Model}, \textsc{sam}, which attends between the dialog history, $c = {c_1, \dots, c_N}$ and the schema graph. \textsc{sam} extends the schema-guided model presented by \citet{mosig2020star} by (1) leveraging a stronger attention mechanism, (2) improving the training algorithm, and (3) removing the linear classification layer which is detrimental to zero-shot performance.

The objective of \textsc{sam} is to predict the node in the schema graph that best corresponds to the dialog context. \textsc{sam} will produce a probability distribution over the nodes corresponding to user utterances and database responses. Given an attention distribution over the nodes, we can obtain a probability distribution over the set of actions by propagating the attention probabilities over the graph. Concretely, if node $u$ has an attention weight of $p$, we add $p$ to the probability of $\textbf{action}(\textbf{next}(u))$.

We consider every node $u$ that corresponds to either a database response or a user utterance. We then represent each node $u$ as the concatenation of the previous node and the current node, i.e., $\textbf{text}(\textbf{prev}(u)) + \textbf{text}(u)$. For all nodes $u \in U$, we obtain this textual representation denoted as $s \in S$.

We are given a language encoder, $\mathcal{F}$, the dialog context, $c = {c_1, \dots, c_N}$, the nodes $U$, their corresponding textual representations $S$, and the set of possible actions $A$. Note that unlike in Equation 1, $\mathcal{F}$ is used to produce a vector representation of each word in the input. \textsc{sam} produces a probability distribution over the actions as follows:

\begin{gather}
    \vh_{1,\dots,N} = \mathcal{F}(c \colon c_1, \dots, c_N) \\
    \mS_{i ; 1,\dots,M} = \mathcal{F}(S_i \colon s_1,\dots,s_M) \\
    \vw^{i}_{j,k} = \vh_j^T  \mS_{i;k} \\  
    \mathbf{\alpha} = \text{softmax}(\vw^{1,\dots,|S|})\\
    p_i = \sum_{j \leq N} \sum_{k \leq M} \alpha^i_{j,k}
\end{gather}

Here, $\vw^i$ is an $N \times M$ dimensional matrix corresponding to the dot product between the $N$ words of the dialog history and the $M$ words of the $i$-th textual representation in $S$. To get the attention weights over all of the words of the schema, we perform a softmax over all $\vw^i, 1 \leq i \leq |S|$. By summing over the attention weights in $\alpha^i$, we get $p_i$, a scalar value which denotes the attention between the dialog history and the $i$-th node (i.e., the corresponding textual representation $S_i$). Given $p_i$ we produce a probability distribution over the actions $A$ as follows:

\begin{gather}
    g(i,a) = \begin{cases}
        p_i, & \text{if } \textbf{action}(\textbf{next}(u_i)) = a\\
        0, & \text{otherwise}
    \end{cases}\\
    P(a) = \sum_{i \leq |S|} g(i,a)
\end{gather}

To align the dialog history to the schema graph, \textsc{sam} performs word-level attention using a BERT-base model. In contrast, the schema-guided model of \citet{mosig2020star} attends with the sentence level vector representation produced by BERT. With the word-level attention, \textsc{sam} can better align ambiguous dialog contexts, such as situations where the user provides multiple pieces of information in a single utterance. Since this word-level attention operates on the sub-word tokens used in BERT, it can also potentially handle spelling errors in the user utterances. 

Furthermore, in their schema-guided model, \citet{mosig2020star} combine the probability distribution produced by attending to the schema graph with their baseline model (i.e., Section 3.1). While this may result in better performance on the tasks the model is trained with, the baseline model will not generalize to unseen tasks. In contrast, \textsc{sam} computes the probability for an action using only the attention over the schema graph. 

\citet{mosig2020star} train their schema-guided model to predict the appropriate node, $u_i$, from a set of nodes $U'$ (s.t., $U' \subset U$). At training time, for efficiency reasons, the set of nodes $U'$ is obtained by using the corresponding node for every dialog context in the training batch. Since the training batches are randomly sampled, this results in $U'$ including nodes from a variety of different schema graphs. At inference time, the dialog task is known and therefore only the corresponding schema graph needs to be attended to (i.e., $U'$ will contain nodes from a single schema graph). It is valuable to train the model to distinguish between different nodes of the same schema graph. Specifically, the attention mechanism (i.e., Equations 5 - 6) will learn stronger fine-grained relationships when trained with negative samples from the \textit{same} domain. As such, we augment the training algorithm to sample batches from the same dialog task, meaning that $U'$ will only include nodes from a single schema. 

\textsc{sam} improves on the baseline schema-guided model introduced by \citet{mosig2020star} by (1) leveraging a stronger attention mechanism that better handles realistic user behavior, (2) computing a probability distribution \textit{only} by attending to the schema graph and (3) modifying the training algorithm to have in-domain negative samples which result in the model learning to identify fine-grained relationships. In combination with the improved schema representation, \textsc{sam} is better suited to handle realistic user behavior in zero-shot settings. 
\begin{table}[]
\renewcommand*{\arraystretch}{1.1}

    \centering
{%
\begin{tabularx}{0.4\textwidth}{@{\extracolsep{\stretch{1}}} c *{2}{c}}
\toprule
\textbf{Model} & \textbf{$F_1$ score}  & \textbf{Accuracy}  \\ \midrule
Baseline $\diamond$  & \textbf{73.79} & \textbf{74.85}  \\
\textsc{bert+s} $\diamond$ & 71.59 & 72.27 \\
\textsc{sam} $-$ [1] & 54.35 & 60.51 \\
\textsc{sam} $-$ [2,3,4] & 70.22 & 71.01 \\
\textsc{sam} $-$ [2] & 70.27 & 71.93  \\
\textsc{sam} $-$ [3] & 70.18 & 71.64 \\
\textsc{sam} $-$ [4] & 69.68 & 69.79 \\
\textsc{sam} & 70.38 & 71.45 \\
\bottomrule
\end{tabularx}
}
    \caption{Performance in the standard experimental setting. Models marked with $\diamond$ are attributed to \citet{mosig2020star}. We denote their schema-guided model, `BERT + \textit{Schema}', as \textit{\textsc{bert+s}}. \textsc{sam} consists of four improvements upon \textsc{bert+s}: (1) user-aware schema, (2) word-level attention, (3) using negative samples from the same task at training, (4) removing the linear classification layer. Results in boldface are statistically significant by t-test ($p < 0.01$)}
    \label{tab:standard}
\end{table}

\begin{table*}[]
\renewcommand*{\arraystretch}{1.1}

    \centering
{%
\begin{tabularx}{0.8\textwidth}{@{\extracolsep{\stretch{1}}} c *{4}{c}}
\toprule
\textbf{Model} & \multicolumn{2}{c}{\textbf{Task Transfer}} & \multicolumn{2}{c}{\textbf{Domain Transfer}}  \\
& \textbf{$F_1$ score} & \textbf{Accuracy} & \textbf{$F_1$ score} & \textbf{Accuracy} \\
\midrule
Baseline $\diamondsuit$  & 31.23 & 30.65 & 31.82 & 33.92 \\
\textsc{bert+s} $\diamondsuit$ & 28.12 & 28.28 & 29.70  & 32.43\\
\textsc{sam} $-$ [1] & 33.81 & 37.84 & 41.77 & 45.64  \\
\textsc{sam} $-$ [2,3,4] & 43.28 & 46.11 & 43.78 & 45.19 \\
\textsc{sam} $-$ [2] & 50.72 & 53.69 & 52.20  & 54.68 \\
\textsc{sam} $-$ [3] & 45.54 & 49.29 & 50.56 & 52.13 \\
\textsc{sam} $-$ [4] & 47.26 & 47.99 & 47.67 & 48.92 \\
\textsc{sam} & \textbf{53.31} & \textbf{55.51} & \textbf{55.74} & \textbf{57.75}\\
\bottomrule
\end{tabularx}
}
    \caption{Performance in zero-shot transfer. We present results on both task transfer and domain transfer. Models marked with $\diamondsuit$ are attributed to \citet{mosig2020star}. \textsc{sam} consists of four improvements upon \textsc{bert+s}: (1) user-aware schema, (2) word-level attention, (3) using negative samples from the same task at training, (4) removing the linear classification layer. Results in bold-face are statistically significant by t-test ($p < 0.01$).}
    \label{tab:zero}
\end{table*}

\section{Results}

To validate the effectiveness of \textsc{sam}, a number of \textit{next action prediction} experiments are carried out on the STAR dataset \citep{mosig2020star}. First, \textsc{sam} is evaluated in the standard experimental setting, i.e., training and testing on the same tasks. Next, we carry out zero-shot transfer experiments, as defined in Section 3. The evaluation uses accuracy and weighted $F_1$ score.

We rerun the experiments presented by \citet{mosig2020star} using code shared by the authors. In our results, the model introduced by \citet{mosig2020star} is denoted as \textsc{bert+s}. Their original results were obtained on an older version of STAR, with annotation errors\footnote{Specifically, certain dialogs were misattributed as being \textit{happy} single-task dialogs.} that have since been fixed. 

\subsection{Standard Experiments}

In the standard experimental setting, models are trained and tested on the same tasks. Following \citet{mosig2020star}, 80\% of the dialogs are used for training and 20\% for testing. All models are trained for 50 epochs. 

The results shown in Table \ref{tab:standard} show \textsc{sam} to be comparable to the baseline model on the standard setting. Since the augmentations to \textsc{sam} are primarily intended to improve zero-shot performance, it is unsurprising that there is no performance improvement compared to the standard setting. When evaluating on \textit{seen} tasks, the linear classification layer is significantly more effective than attending to the schema. This suggests that a large neural model (i.e., BERT) is able to implicitly learn meaningful dialog policies from dialog data. It is possible that this performance difference may decrease with more expressive schemas (e.g., having multiple examples for each user utterance, automatically learning schemas from the dataset). The value of our schema graphs is nonetheless shown when comparing \textsc{sam} to \textsc{sam}$-[1]$ (i.e., the old schema graphs). These experiments provide an upper bound for the performance in zero-shot transfer.

\subsection{Zero-Shot Transfer}

Table \ref{tab:zero} shows the results of the zero-shot experiments. \textsc{sam} obtains strong improvements over the baseline models for both zero-shot task transfer and domain transfer. These experimental results validate the effectiveness of the schema-guided paradigm, as well as the specific design of \textsc{sam}.

Compared to the baseline model (described in Section 3.1), \textsc{sam} obtains a $\mathbf{+22 ~F_1}$ score improvement in task transfer and a $\mathbf{+24 ~F_1}$ score improvement in domain transfer. Since the baseline model is unable to predict classes it has not observed at training time, its performance is limited to actions that are consistent across domains (e.g., \textit{`hello'}, \textit{`goodbye'}, \textit{`anything-else'}). This improvement highlights the effectiveness of the schema-guided paradigm for zero-shot transfer learning.

\textsc{bert+s} also leverages schemas for transfer learning. Yet, it under-performs relative to the baseline model. \textsc{sam} attains even larger improvements over this baseline schema-guided model. As described in Section 4.2, the weak performance of \textsc{bert+s} is largely a consequence of it being incapable of handling realistic user behavior. The design of \textsc{bert+s} (i.e., the schema only having system nodes) results in the model essentially predicting the subsequent system actions. This is equivalent to sequentially predicting the next system action, regardless of user behavior. With improved schema representations and model architecture, \textsc{sam} achieves much stronger performance in zero-shot transfer.

Our ablation experiments shed more light on the performance of \textsc{sam} relative to \textsc{bert+s}. A significant performance drop is observed when removing the newly constructed schema representations (i.e., \textsc{sam}$- [1]$). In contrast, adding the schema graphs to \textsc{bert+s} (i.e., \textsc{sam}$- [2,3,4]$) results in a strong performance improvement of $\mathbf{+15~F_1}$ score. This confirms the hypothesis that the schema graphs of \citet{mosig2020star}, which are largely comprised of system action nodes are insufficient for modelling realistic user behavior. 

Word-level attention is shown to give moderate, albeit statistically significant, improvement. In contrast to \textsc{sam}$-[2]$, \textsc{sam} obtains a $\mathbf{+3~F_1}$ score improvement. While word-level attention allows the model to better align the dialog to the schema, it is an architectural improvement that is not central to the schema-guided paradigm.

Modifying the training algorithm to sample batches from the same task results in better negative samples during training. This allows the model to learn to distinguish between nodes from the same schema graph when aligning the dialog to the schema graph. When this modification is removed (i.e., \textsc{sam}$-[3])$, the performance of \textsc{sam} drops by $\mathbf{8~F_1}$ score for zero-shot task transfer.

The fourth and final component of \textsc{sam} is the removal of the linear classification layer. Since this classification layer is unable to predict classes it has not seen at training time, it is ineffective in zero-shot settings. Unsurprisingly, removing it increases performance and \textsc{sam} obtains a $\mathbf{+6~F_1}$ score improvement over \textsc{sam}$-[4]$.

The zero-shot experiments shown in Table \ref{tab:zero} empirically validate several hypotheses presented in this paper. First, the strong improvement over the baseline demonstrates the efficacy of the schema-guided paradigm for zero-shot generalizability in end-to-end dialog. Decoupling dialog policy and the language understanding by explicitly representing the task-specific dialog policies as schema graphs results in an improved ability to transfer to unseen tasks. Next, we improve over the schema-guided model of \citet{mosig2020star} through (1) an improved schema representation and (2) a collection of modifications to the model. The improved schema representation better models realistic user behaviors in dialog, and therefore results in better alignment of the dialog and the schema. Our model modifications result in the model being able to learn better fine-grained relationships during alignment (e.g., through better negative sampling and word-level attention) and better handle zero-shot transfer (e.g., by removing the linear layer).

In contrast to prior work on zero-shot generalizability \citep{zhao2018zero,qian2019domain}, our approach is shown to effectively  transfer between the vastly dissimilar domains of the STAR corpus \citep{mosig2020star} (e.g., trivia or spaceship maintenance). Rather than modelling a cross-domain mapping and leveraging similar concepts across different domains, the schema-guided paradigm \textit{decouples} the domain-specific (i.e., the dialog policy) and domain-agnostic (i.e., language understanding) aspects of dialog systems. Through the schema-guided paradigm, we achieve strong performance in the zero-shot setting and take an important step towards zero-shot dialog.

\section{Conclusion}

This paper shows strong results in zero-shot task transfer and domain transfer using the schema-guided paradigm. We hypothesized that the difficulty of zero-shot transfer in dialog stems from the dialog policy. When neural models implicitly memorize dialog policies observed at training time, they struggle to transfer to new tasks. To mitigate this, we explicitly provide the dialog policy to the model, in the form of a schema graph. This paper introduces the Schema Attention Model (\textsc{sam}) and shows improved schema graphs for the STAR corpus. This approach attains significant improvement over prior work in the zero-shot setting, with a \textbf{$\mathbf{+22 ~F_1}$ score improvement}. Furthermore, the ablation experiments demonstrate the effectiveness of both \textsc{sam} and the improved schema representations. Future work may explore (1) improved schema representations to better capture dialog policy, (2) improved model architectures to better align the dialog to the schema, and (3) extensions to other problems (e.g., response generation).
\bibliographystyle{acl_natbib}
\bibliography{anthology,acl2021}

\begin{thebibliography}{30}
\expandafter\ifx\csname natexlab\endcsname\relax\def\natexlab#1{#1}\fi

\bibitem[{Adiwardana et~al.(2020)Adiwardana, Luong, So, Hall, Fiedel,
  Thoppilan, Yang, Kulshreshtha, Nemade, Lu et~al.}]{adiwardana2020towards}
Daniel Adiwardana, Minh-Thang Luong, David~R So, Jamie Hall, Noah Fiedel, Romal
  Thoppilan, Zi~Yang, Apoorv Kulshreshtha, Gaurav Nemade, Yifeng Lu, et~al.
  2020.
\newblock Towards a human-like open-domain chatbot.
\newblock \emph{arXiv preprint arXiv:2001.09977}.

\bibitem[{Bapna et~al.(2017)Bapna, Tur, Hakkani-Tur, and
  Heck}]{bapna2017towards}
Ankur Bapna, Gokhan Tur, Dilek Hakkani-Tur, and Larry Heck. 2017.
\newblock Towards zero-shot frame semantic parsing for domain scaling.
\newblock \emph{arXiv preprint arXiv:1707.02363}.

\bibitem[{Bohus and Rudnicky(2009)}]{bohus2009ravenclaw}
Dan Bohus and Alexander~I Rudnicky. 2009.
\newblock The ravenclaw dialog management framework: Architecture and systems.
\newblock \emph{Computer Speech \& Language}, 23(3):332--361.

\bibitem[{Budzianowski et~al.(2018)Budzianowski, Wen, Tseng, Casanueva, Ultes,
  Ramadan, and Ga{\v{s}}i{\'c}}]{budzianowski2018multiwoz}
Pawe{\l} Budzianowski, Tsung-Hsien Wen, Bo-Hsiang Tseng, I{\~n}igo Casanueva,
  Stefan Ultes, Osman Ramadan, and Milica Ga{\v{s}}i{\'c}. 2018.
\newblock Multiwoz-a large-scale multi-domain wizard-of-oz dataset for
  task-oriented dialogue modelling.
\newblock \emph{arXiv preprint arXiv:1810.00278}.

\bibitem[{Chen et~al.(2016)Chen, Hakkani-T{\"u}r, and He}]{chen2016zero}
Yun-Nung Chen, Dilek Hakkani-T{\"u}r, and Xiaodong He. 2016.
\newblock Zero-shot learning of intent embeddings for expansion by
  convolutional deep structured semantic models.
\newblock In \emph{2016 IEEE International Conference on Acoustics, Speech and
  Signal Processing (ICASSP)}, pages 6045--6049. IEEE.

\bibitem[{Devlin et~al.(2019)Devlin, Chang, Lee, and
  Toutanova}]{devlin-etal-2019-bert}
Jacob Devlin, Ming-Wei Chang, Kenton Lee, and Kristina Toutanova. 2019.
\newblock \href {https://doi.org/10.18653/v1/N19-1423} {{BERT}: Pre-training of
  deep bidirectional transformers for language understanding}.
\newblock In \emph{Proceedings of the 2019 Conference of the North {A}merican
  Chapter of the Association for Computational Linguistics: Human Language
  Technologies, Volume 1 (Long and Short Papers)}, pages 4171--4186,
  Minneapolis, Minnesota. Association for Computational Linguistics.

\bibitem[{Ferguson and Allen(1998)}]{ferguson1998trips}
George Ferguson and James Allen. 1998.
\newblock Trips: An intelligent integrated problem-solving assistant.
\newblock In \emph{Proceedings of the Fifteenth National Conference on
  Artificial Intelligence (AAAI-98)}, pages 567--573.

\bibitem[{Hemphill et~al.(1990)Hemphill, Godfrey, and
  Doddington}]{hemphill-etal-1990-atis}
Charles~T. Hemphill, John~J. Godfrey, and George~R. Doddington. 1990.
\newblock \href {https://www.aclweb.org/anthology/H90-1021} {The {ATIS} spoken
  language systems pilot corpus}.
\newblock In \emph{Speech and Natural Language: Proceedings of a Workshop Held
  at Hidden Valley, {P}ennsylvania, June 24-27,1990}.

\bibitem[{Hu et~al.(2019)Hu, Chan, Liu, Zhao, Ma, and Yan}]{hu2019gsn}
Wenpeng Hu, Zhangming Chan, Bing Liu, Dongyan Zhao, Jinwen Ma, and Rui Yan.
  2019.
\newblock Gsn: A graph-structured network for multi-party dialogues.
\newblock \emph{arXiv preprint arXiv:1905.13637}.

\bibitem[{Jurafsky(2000)}]{jurafsky2000speech}
Dan Jurafsky. 2000.
\newblock \emph{Speech \& language processing}.
\newblock Pearson Education India.

\bibitem[{Mehri et~al.(2020)Mehri, Eric, and Hakkani-Tur}]{mehri2020dialoglue}
Shikib Mehri, Mihail Eric, and Dilek Hakkani-Tur. 2020.
\newblock Dialoglue: A natural language understanding benchmark for
  task-oriented dialogue.
\newblock \emph{arXiv preprint arXiv:2009.13570}.

\bibitem[{Mosig et~al.(2020)Mosig, Mehri, and Kober}]{mosig2020star}
Johannes~EM Mosig, Shikib Mehri, and Thomas Kober. 2020.
\newblock Star: A schema-guided dialog dataset for transfer learning.
\newblock \emph{arXiv preprint arXiv:2010.11853}.

\bibitem[{Qian and Yu(2019)}]{qian2019domain}
Kun Qian and Zhou Yu. 2019.
\newblock Domain adaptive dialog generation via meta learning.
\newblock \emph{arXiv preprint arXiv:1906.03520}.

\bibitem[{Qiu et~al.(2020)Qiu, Zhao, Shi, Liang, Shi, Yuan, Yu, and
  Zhu}]{qiu2020structured}
Liang Qiu, Yizhou Zhao, Weiyan Shi, Yuan Liang, Feng Shi, Tao Yuan, Zhou Yu,
  and Song-Chun Zhu. 2020.
\newblock Structured attention for unsupervised dialogue structure induction.
\newblock \emph{arXiv preprint arXiv:2009.08552}.

\bibitem[{Radford et~al.(2019)Radford, Wu, Child, Luan, Amodei, and
  Sutskever}]{radford2019language}
Alec Radford, Jeffrey Wu, Rewon Child, David Luan, Dario Amodei, and Ilya
  Sutskever. 2019.
\newblock Language models are unsupervised multitask learners.
\newblock \emph{OpenAI blog}, 1(8):9.

\bibitem[{Rastogi et~al.(2020{\natexlab{a}})Rastogi, Zang, Sunkara, Gupta, and
  Khaitan}]{rastogi2020schema}
Abhinav Rastogi, Xiaoxue Zang, Srinivas Sunkara, Raghav Gupta, and Pranav
  Khaitan. 2020{\natexlab{a}}.
\newblock Schema-guided dialogue state tracking task at dstc8.
\newblock \emph{arXiv preprint arXiv:2002.01359}.

\bibitem[{Rastogi et~al.(2020{\natexlab{b}})Rastogi, Zang, Sunkara, Gupta, and
  Khaitan}]{rastogi2020towards}
Abhinav Rastogi, Xiaoxue Zang, Srinivas Sunkara, Raghav Gupta, and Pranav
  Khaitan. 2020{\natexlab{b}}.
\newblock Towards scalable multi-domain conversational agents: The
  schema-guided dialogue dataset.
\newblock In \emph{Proceedings of the AAAI Conference on Artificial
  Intelligence}, volume~34, pages 8689--8696.

\bibitem[{Raux et~al.(2005)Raux, Langner, Bohus, Black, and
  Eskenazi}]{raux2005let}
Antoine Raux, Brian Langner, Dan Bohus, Alan~W Black, and Maxine Eskenazi.
  2005.
\newblock Let's go public! taking a spoken dialog system to the real world.
\newblock In \emph{Ninth European conference on speech communication and
  technology}.

\bibitem[{Rich and Sidner(1998)}]{rich1998collagen}
Charles Rich and Candace~L Sidner. 1998.
\newblock Collagen: A collaboration manager for software interface agents.
\newblock In \emph{Computational models of mixed-initiative interaction}, pages
  149--184. Springer.

\bibitem[{Shah et~al.(2019)Shah, Gupta, Fayazi, and
  Hakkani-Tur}]{shah2019robust}
Darsh~J Shah, Raghav Gupta, Amir~A Fayazi, and Dilek Hakkani-Tur. 2019.
\newblock Robust zero-shot cross-domain slot filling with example values.
\newblock \emph{arXiv preprint arXiv:1906.06870}.

\bibitem[{Shi et~al.(2019)Shi, Zhao, and Yu}]{shi2019unsupervised}
Weiyan Shi, Tiancheng Zhao, and Zhou Yu. 2019.
\newblock Unsupervised dialog structure learning.
\newblock \emph{arXiv preprint arXiv:1904.03736}.

\bibitem[{Wang et~al.(2019)Wang, Pruksachatkun, Nangia, Singh, Michael, Hill,
  Levy, and Bowman}]{wang2019superglue}
Alex Wang, Yada Pruksachatkun, Nikita Nangia, Amanpreet Singh, Julian Michael,
  Felix Hill, Omer Levy, and Samuel~R Bowman. 2019.
\newblock Superglue: A stickier benchmark for general-purpose language
  understanding systems.
\newblock \emph{arXiv preprint arXiv:1905.00537}.

\bibitem[{Wang et~al.(2018)Wang, Singh, Michael, Hill, Levy, and
  Bowman}]{wang2018glue}
Alex Wang, Amanpreet Singh, Julian Michael, Felix Hill, Omer Levy, and Samuel~R
  Bowman. 2018.
\newblock Glue: A multi-task benchmark and analysis platform for natural
  language understanding.
\newblock \emph{arXiv preprint arXiv:1804.07461}.

\bibitem[{Wen et~al.(2016)Wen, Vandyke, Mrksic, Gasic, Rojas-Barahona, Su,
  Ultes, and Young}]{wen2016network}
Tsung-Hsien Wen, David Vandyke, Nikola Mrksic, Milica Gasic, Lina~M
  Rojas-Barahona, Pei-Hao Su, Stefan Ultes, and Steve Young. 2016.
\newblock A network-based end-to-end trainable task-oriented dialogue system.
\newblock \emph{arXiv preprint arXiv:1604.04562}.

\bibitem[{Williams and Zweig(2016)}]{williams2016end}
Jason~D Williams and Geoffrey Zweig. 2016.
\newblock End-to-end lstm-based dialog control optimized with supervised and
  reinforcement learning.
\newblock \emph{arXiv preprint arXiv:1606.01269}.

\bibitem[{Wu et~al.(2019)Wu, Madotto, Hosseini-Asl, Xiong, Socher, and
  Fung}]{wu2019transferable}
Chien-Sheng Wu, Andrea Madotto, Ehsan Hosseini-Asl, Caiming Xiong, Richard
  Socher, and Pascale Fung. 2019.
\newblock Transferable multi-domain state generator for task-oriented dialogue
  systems.
\newblock \emph{arXiv preprint arXiv:1905.08743}.

\bibitem[{Xu et~al.(2020)Xu, Lei, Wang, Niu, Wu, Che, and
  Liu}]{xu2020discovering}
Jun Xu, Zeyang Lei, Haifeng Wang, Zheng-Yu Niu, Hua Wu, Wanxiang Che, and Ting
  Liu. 2020.
\newblock Discovering dialog structure graph for open-domain dialog generation.
\newblock \emph{arXiv preprint arXiv:2012.15543}.

\bibitem[{Zhang et~al.(2020)Zhang, Sun, Galley, Chen, Brockett, Gao, Gao, Liu,
  and Dolan}]{zhang-etal-2020-dialogpt}
Yizhe Zhang, Siqi Sun, Michel Galley, Yen-Chun Chen, Chris Brockett, Xiang Gao,
  Jianfeng Gao, Jingjing Liu, and Bill Dolan. 2020.
\newblock \href {https://doi.org/10.18653/v1/2020.acl-demos.30} {{DIALOGPT} :
  Large-scale generative pre-training for conversational response generation}.
\newblock In \emph{Proceedings of the 58th Annual Meeting of the Association
  for Computational Linguistics: System Demonstrations}, pages 270--278,
  Online. Association for Computational Linguistics.

\bibitem[{Zhao and Eskenazi(2018)}]{zhao2018zero}
Tiancheng Zhao and Maxine Eskenazi. 2018.
\newblock Zero-shot dialog generation with cross-domain latent actions.
\newblock \emph{arXiv preprint arXiv:1805.04803}.

\bibitem[{Zhao et~al.(2017)Zhao, Lu, Lee, and Eskenazi}]{zhao2017generative}
Tiancheng Zhao, Allen Lu, Kyusong Lee, and Maxine Eskenazi. 2017.
\newblock Generative encoder-decoder models for task-oriented spoken dialog
  systems with chatting capability.
\newblock \emph{arXiv preprint arXiv:1706.08476}.

\end{thebibliography}

\clearpage

\appendix

\section{Example Outputs}

We show a few example outputs in Table \ref{tab:examples}. These examples are model outputs produced in the zero-shot domain transfer experiments. None of these models have seen any training data pertaining to the specific domains. These examples demonstrate the effectiveness of \textsc{sam} to handle ambiguous user intents and effectively align them to the dialog schema.

\begin{table*}[]
\renewcommand*{\arraystretch}{1.1}

{%
\begin{tabularx}{\textwidth}{@{\extracolsep{\stretch{1}}} l p{0.8\textwidth}}
\toprule

\textbf{Domain:} & Spaceship \\
\textbf{Task:} & Spaceship Access Codes \\
\textbf{Dialog History:} & \textcolor{blue}{USER:} The ship is under attack! Need help ASAP \textcolor{blue}{SYSTEM:} Please provide your name. \textcolor{blue}{USER:} Rey. I'm the first officer \textcolor{blue}{SYSTEM:} Please enter the code. \textcolor{blue}{USER:} I think it's 9315831 \\

\textbf{Baseline Output:} & \textcolor{red}{Query} \\
\textbf{\textsc{bert+s} Output:} & \textcolor{red}{Query} \\
\textbf{\textsc{sam} Output:} & \textcolor{green}{Spaceship Ask Code Type} (\textit{`Please specify the code type.'}) \\
\textbf{\textsc{sam} Schema Node:} & \textcolor{blue}{SYSTEM:} Please enter the code. \textcolor{blue}{USER:} [NUMBER] \\

\midrule

\textbf{Domain:} & Bank \\
\textbf{Task:} & Check Bank Balance \\

\textbf{Dialog History:} & \textcolor{blue}{USER:}  Hi, can you help me check my balance? \textcolor{blue}{SYSTEM:} Could I get your full name, please? \textcolor{blue}{USER:} Jane Doe \textcolor{blue}{SYSTEM:} Could you tell me your account number, please? \textcolor{blue}{USER:} I don't remember it or my PIN number unfortunately. \\

\textbf{Baseline Output:} & \textcolor{red}{Query} \\
\textbf{\textsc{bert+s} Output:} & \textcolor{red}{Inform Cannot Authenticate} (\textit{`I am sorry, but I cannot authenticate you with the information you have provided.'})\\
\textbf{\textsc{sam} Output:} & \textcolor{green}{Ask Date of Birth} (\textit{`Could you provide your date of birth, please?'}) \\
\textbf{\textsc{sam} Schema Node:} & \textcolor{blue}{SYSTEM:} Could you tell me your account number, please?  \textcolor{blue}{USER:} I don't know my account number. \\

\bottomrule

\end{tabularx}
}
    \caption{Example outputs in the zero-shot domain transfer setting. The models have not seen any data pertaining to the domains that are tested here. The first example shows that \textsc{sam} can effectively align to the schema. The second example is more complicated and requires \textsc{sam} to identify that the user indicated that they don't remember their account number (note that the most typical system response following \textit{`Can you tell me your account number, please?')} is asking for the user's PIN.}
    \label{tab:examples}
\end{table*}

\section{Example Schemas}

All of the schema graphs produced for this paper are available in our GitHub Repository\footnote{\url{https://github.com/Shikib/schema_attention_model}}.  These schemas improve upon the schemas provided by \citet{mosig2020star}. Nonetheless, this schema representation is still far from perfect. Future work may explore designing more expressive and thorough schemas. There are several approaches for doing this: (1) having multiple examples for each user utterance, (2) better representing loops in the schema graphs (e.g., query the database until you find the cheapest flight), (3) representing more sophisticated (potentially dynamic) scenarios (e.g., if the user's answer matches the correct answer to a trivia question, inform them that they are correct). The proposed schema representations address several weaknesses, however there is still significant room for improvement.

\end{document}